\documentclass{article}
\usepackage{spconf,amsmath,graphicx}
\usepackage{multirow}
\usepackage{booktabs}
\usepackage{xcolor}
\usepackage{hyperref}  % 建议放最后加载

\graphicspath{{figures/}} % 把所有图片放到 figures/ 目录下（非常重要！）

\title{MedSpeak: A Knowledge Graph-Aided ASR Error Correction Framework for Spoken Medical QA}

\name{
\shortstack{
Yutong Song$^{\star}$, Shiva Shrestha$^{\dagger}$, Chenhan Lyu$^{\star}$, Elahe Khatibi$^{\star}$,\\
Pengfei Zhang$^{\star}$, Honghui Xu$^{\dagger}$, Nikil Dutt$^{\star}$, Amir Rahmani$^{\star}$
}}
\address{
$^{\star}$University of California, Irvine, Computer Science, Irvine, CA, USA\\
$^{\dagger}$Kennesaw State University, Computer Science, Kennesaw, GA, USA}

\begin{document}
\maketitle

\begin{abstract}
Spoken question-answering (SQA) systems relying on automatic speech recognition (ASR) often struggle with accurately recognizing medical terminology.
To this end, we propose MedSpeak, a novel knowledge graph-aided ASR error correction framework that refines noisy transcripts and improves downstream answer prediction by leveraging both semantic relationships and phonetic information encoded in a medical knowledge graph, together with the reasoning power of LLMs.
Comprehensive experimental results on benchmarks demonstrate that MedSpeak significantly improves the accuracy of medical term recognition and overall medical SQA performance, establishing MedSpeak as a state-of-the-art solution for medical SQA. The code is available at \url{https://github.com/RainieLLM/MedSpeak}.


\end{abstract}

%\begingroup\def\thefootnote{}\footnotetext{The code is available at \url{https://github.com/RainieLLM/MedSpeak}}\endgroup.

%===================================================================
\section{Introduction}
The increasing adoption of speech-based AI systems in medical applications has led to the growing need for high-accuracy spoken medical question answering (SQA) systems. These systems rely on Automatic Speech Recognition (ASR) to transcribe spoken medical queries before feeding them into retrieval-augmented generation (RAG) models or large language models (LLMs) for answer generation. However, despite advancements in ASR models, such as Whisper and Wav2Vec2, their performance remains suboptimal in medical settings due to several domain-specific challenges~\cite{zero-shot-sqa, gec-rag}. These challenges include complex medical terminology, phonetic ambiguities, and contextual dependencies, which degrade ASR transcription accuracy and impact downstream medical reasoning~\cite{retrieval-augmented-named-entity, retrieval-augmented-end-to-end}. 
In other words, traditional ASR models, trained on general-domain corpora, often struggle to accurately recognize specialized medical terminology, leading to frequent misrecognitions of critical medical entities~\cite{ retrieval-augmented-speech-rag, medrag-healthcare}.
Although some domain-specific finetuning approaches have been attempted~\cite{gec-rag, retrieval-augmented-slu, seal-rag},
%but is 
they are
restricted by data insufficiency and lack of generalization. Additionally, these domain-specific models still face challenges with phonetic confusability between terms like "hypertension" and "hypotension," further exacerbating ASR errors~\cite{retrieval-augmented-text-to-audio}. 
While distinct in their medical implications, those words can be confusingly similar to ASR systems, leading to errors in transcription. 
These deficiencies have negative impacts in clinical applications where transcription errors cascade to incorrect reasoning with resulting threats to patient safety. 
This underscores the need for a robust ASR with a reliable QA system to support trustworthy healthcare decisions.
\begin{figure}[t]
    \centering
    \includegraphics[width=0.4\textwidth]{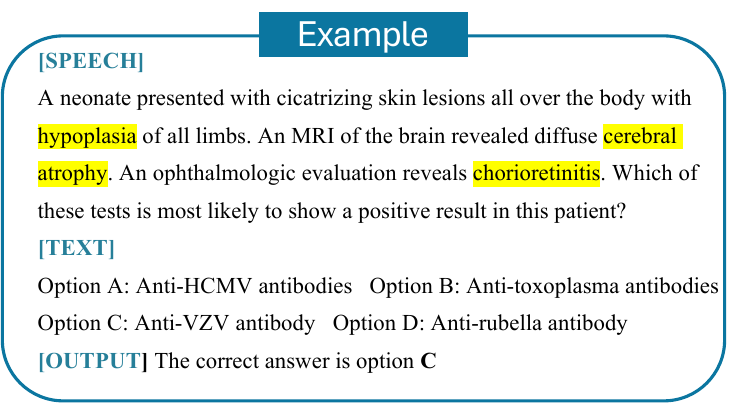} 
    \vspace{-1em}
    \caption{An Example of SQA.}
    \label{fig:example}
    \vspace{-2em}
\end{figure}
 
To address these challenges, in this paper, we propose MedSpeak, a knowledge graph-aided ASR error correction framework designed specifically for spoken medical QA tasks. By harnessing the power of knowledge graphs and the reasoning capabilities of LLMs, MedSpeak integrates structured domain knowledge, phonetic and semantic similarity constraints, and context-aware reasoning to significantly improve medical ASR correction and QA ability, as shown in Fig.~\ref{fig:framework}. Our contributions include:
(1) An automated medical knowledge graph (KG) construction method encodes semantic and phonetic relationships between terms, enabling robust error correction and reasoning.
(2) A multi-constraint retrieval mechanism leverages phonetic and semantic features to generate accurate ASR hypotheses for medical terminology.
(3) A fine-tuned LLM integrates retrieved knowledge with answer constraints, producing precise transcriptions and reliable answers.
(4) Extensive experiments on spoken medical QA datasets show that MedSpeak achieves state-of-the-art ASR error correction in the medical domain.

%===================================================================
\section{Related Works}
Traditional ASR models, trained on general-domain corpora, often struggle to accurately recognize specialized medical terminology, leading to frequent misrecognitions of critical medical entities \cite{zero-shot-sqa, retrieval-augmented-speech-rag, medrag-healthcare, retrieval-augmented-slu}. While fine-tuning ASR models on domain-specific datasets has been explored, these efforts are still limited by data scarcity and poor generalization \cite{gec-rag, retrieval-augmented-slu, seal-rag}. Additionally, these models face challenges with phonetic confusability between terms like "hypertension" and "hypotension," further exacerbating ASR errors~\cite{retrieval-augmented-named-entity, retrieval-augmented-end-to-end, retrieval-augmented-text-to-audio}.

To address these challanges, RAG-based methods aim to improve ASR accuracy and named entity recognition~\cite{lei2024contextualization, pusateri2024retrieval} by incorporating external domain specific knowledge retrieval~\cite{yang2024rasu, xue2024retrieval, xiao2025contextual}. Techniques such as GEC-RAG~\cite{gec-rag} and IRAG \cite{irag} utilize contextually retrieved text snippets to refine ASR hypotheses. However, these approaches relies on text-centric evidences and fail to leverage phonetic knowledge, limiting their ability to resolve confusion between phonetically similar medical entities.  Structured knowledge graphs have been explored for domain-specific ASR correction, particularly in medical QA systems. MedRAG~\cite{medrag} incorporates a medical KG into retrieval-based ASR correction but does not explicitly address phonetic similarity for disambiguation. Similarly, RANEC~\cite{retrieval-augmented-named-entity} leverages named entity correction for ASR refinement but overlooks the integration of multiple-choice context for better accuracy.

%===================================================================
\section{Medspeak}

Our MedSpeak framework combines static KG context injection with a fine-tuned LLM, explicitly integrating both semantic relationship and phonetic similarity into LLM fine-tuning. This combined representation allows for robust ASR error correction and QA reasoning within a unified two-line supervision format, achieving great performance across multiple medical QA benchmarks. Fig.~\ref{fig:example} presents an example of our spoken question-answering (SQA) framework.
\begin{figure}[htbp]
    \centering
    \includegraphics[width=0.5\textwidth]{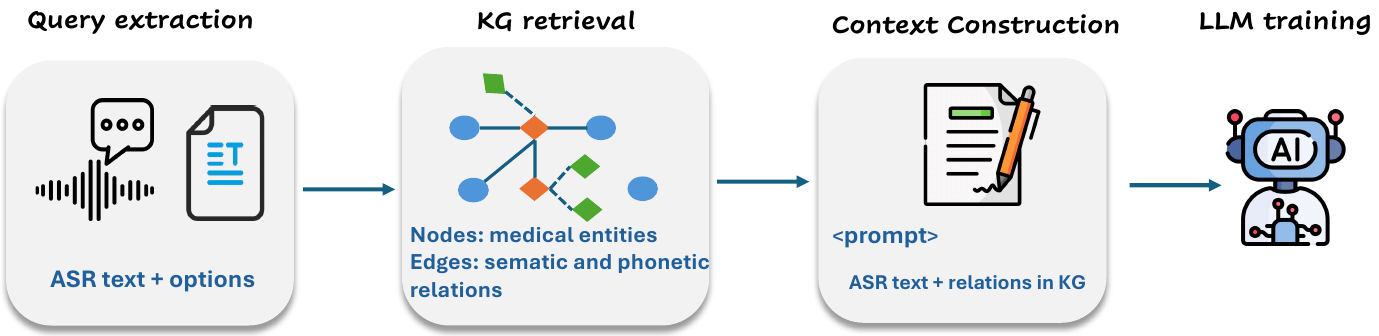}
    \vspace{-1em}
    \caption{Overview of MedSpeak framework.}
    \label{fig:framework}
    \vspace{-1em}
\end{figure}

The system receives two distinct inputs: (1) a clinical case description in speech form (containing key diagnostic terms highlighted in yellow such as "hypoplasia," "cerebral atrophy," and "chorioretinitis"), and (2) four possible diagnostic test options provided as text (Options A-D). The system then produces two outputs: the accurately transcribed speech and the correct option (Option C). This example demonstrates the critical challenge of accurately recognizing specialized medical terminology in speech, as mis-recognition of terms like "chorioretinitis" as "chorioamnionitis" would lead to incorrect evidence retrieval and subsequent diagnostic errors. Our proposed MedSpeak leverages knowledge graph-aided error correction to improve recognition of these critical diagnostic terms, ensuring that the speech-to-text conversion preserves the medical meaning necessary for selecting the appropriate diagnostic test. In the following, we detail two main components in MedSpeak.

%===================================================================
\subsection {Knowledge Graph Building}
% We construct our knowledge graph using the Unified Medical Language System (UMLS) dataset from the National Institutes of Health (NIH) \cite{umls2024}, establishing relationships between medical terms through the MRREL table. During construction, we filter out generic relationships, duplicate entries, and relationship pairs that lack specific semantic information. Many entries in the relationship table contain only relation indicators without detailed descriptions, providing limited value for information extraction. By removing these uninformative relationships, we streamline both the knowledge graph construction process and improve retrieval efficiency.
%
We construct our knowledge graph using the Unified Medical Language System (UMLS) dataset from the National Institutes of Health (NIH) \cite{umls2024}, establishing relationships between medical terms through the MRREL table. During construction, we iteratively select medical keywords from the table while filtering out generic relationships, duplicate entries, and relationship pairs that lack specific semantic information. After each keyword is selected, we use The CMU Pronouncing Dictionary to generate phonetically similar words and append them to our knowledge graph. By removing these uninformative relationships and adding pronunciation information, we streamline both the knowledge graph construction process and improve retrieval efficiency.
\begin{figure}[h]
    \centering
    \includegraphics[width=0.4\textwidth]{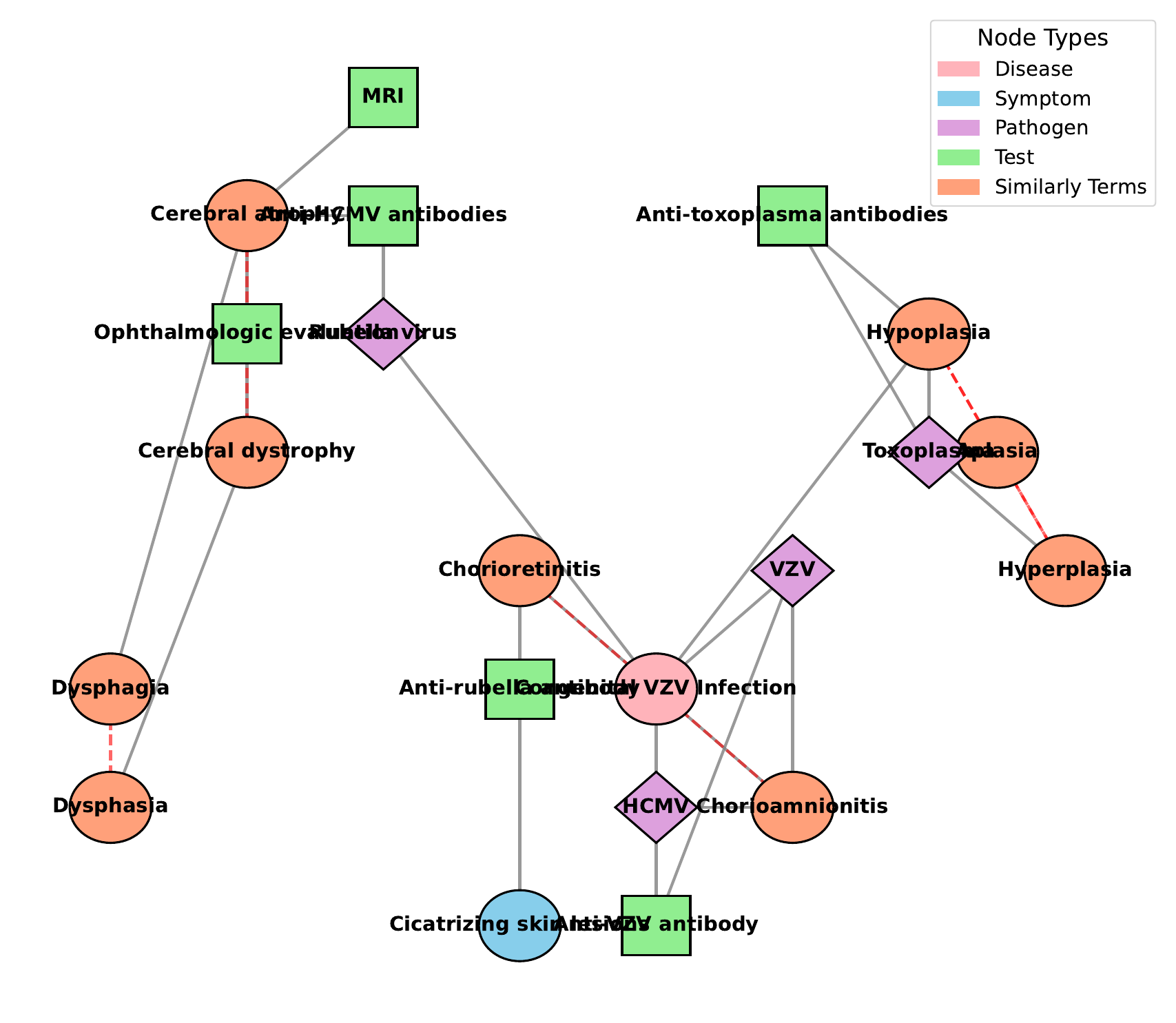} 
    \vspace{-1em}
        \vspace{-1em}
    \caption{Medical KG Constructed from SQA.}
    \label{fig:kg}
\end{figure}

% As outlined in Fig.~\ref{fig:framework}, in the "Relations in the Knowledge Graph" step, we deploy the Double Metaphone algorithm and Levenshtein Distance metric to identify English words with similar pronunciation to medical entities. Since Double Metaphone alone produces excessive false positives, combining both metrics enhances phonetic matching accuracy. We then integrate these entities into a unified knowledge graph for retrieval.
% Fig.\ref{fig:kg} depicts a semantic knowledge graph for the QA example in Fig.\ref{fig:example}. Each node represents differential diagnostic considerations (HCMV, Toxoplasma, VZV, and Rubella virus infections) with corresponding antibody tests. The graph incorporates phonetically similar terms often misrecognized in speech systems, such as "Hypoplasia"/"Hyperplasia" and "atrophy"/"hypertrophy."
% In the final KG, nodes contain unique identifiers and term strings, while edges have relationship tags including "classifies," "constitutes," "due\_to," and "plays\_role" to help the LLM understand relationships between question terms. A specific "phonetic" edge tag aids the LLM in identifying potential speech transcription errors.
As outlined in Fig.~\ref{fig:framework}, in the "Relations in the Knowledge Graph" step, we deploy the Double Metaphone algorithm and Levenshtein Distance metric to identify English words with similar pronunciation to medical entities. Since Double Metaphone alone produces excessive false positives, combining both metrics enhances phonetic matching accuracy. We then integrate these entities into a unified knowledge graph for retrieval.
Fig.\ref{fig:kg} depicts a semantic knowledge graph for the QA example in Fig.\ref{fig:example}. Each node represents differential diagnostic considerations (HCMV, Toxoplasma, VZV, and Rubella virus infections) with corresponding antibody tests. The graph incorporates phonetically similar terms often misrecognized in speech systems, such as "Hypoplasia"/"Hyperplasia" and "atrophy"/"hypertrophy."
In the final KG, nodes contain unique identifiers and term strings, while edges have relationship tags including "classifies," "constitutes," "due\_to," and "plays\_role" to help the LLM understand relationships between question terms. A specific "phonetic" edge tag aids the LLM in identifying potential speech transcription errors.

%===================================================================
\subsection{KG-Aided ASR Error Correction Framework}
\vspace{-0.7em}
In our MedSpeak, the workflow begins with an existing speech recognition model to generate transcripts, followed by identification of medical terminology for knowledge graph retrieval. The retrieved medical information is then embedded alongside contextual prompts to enhance LLM reasoning and correction capabilities.

%=========================================================================
\subsubsection{Training Process}
Our training methodology primarily fine-tunes the parameters of a base LLM model under the full model supervision. Each training sample is formatted as a dialogue with following three parts: \textbf{i) System Messages} consists of a fixed rules only instruction that enforced a strict two-line output from the LLM. \textbf{ii) User Input} included three items:(a) the noisy ASR transcript, (b) the multiple-choice answer options, and (c) the semantic and phonetic Knowledge-Graph (KG). The KG section is truncated to a fixed budget(900 tokens) in order to prevent gold labels from being dropped. Finally, \textbf{iii) Assistant Target} is defined as a structured two-line completion format:

\begin{verbatim}
Corrected Text: [Original GT Text]
Correct Option: [Option Letter<A|B|C|D>]
\end{verbatim}
\noindent
\textbf{Training Objective:} Given an input--output pair $(x,y)$, the model is optimized with a single causal language modeling objective. Formally, the input sequence is defined as $x = \big[ s \,;\, u(\text{ASR}, \text{Options}, \text{KG}) \big]$, where $s$ is the fixed \emph{system instruction}, and $u(\cdot)$ represents the \emph{user input} consisting of the ASR transcript, multiple-choice options, and truncated KG context. The target sequence is expressed as $ y = \textit{Corrected Text: } t^* \oplus \textit{Correct Option: } o^*$, where $t^{*}$ is the \emph{gold-standard transcript (ground truth)} and $o^{*} \in \{A,B,C,D\}$ is the \emph{correct answer option}. The model parameters $\theta$ are trained to minimize
\begin{equation}
    \mathcal{L}(\theta) = - \sum_{i=1}^{|y|} 
    \log P_{\theta}(y_i \mid x, y_{<i}),
\end{equation} 
where $y_i$ is the $i$-th token in the target sequence, $y_{<i}$ are all previous tokens, and $P_\theta(y_i \mid x, y_{<i})$ is the probability of predicting the correct token given the input $x$ and prior context. 
This formulation unifies \emph{ASR error correction} (via $t^*$) and \emph{multiple-choice reasoning} (via $o^*$) into a single generation task. The KG ensures domain knowledge remains available, enabling the model to produce clinically accurate and context-aware outputs.

\subsubsection{Inference Process}
The acoustic input $a$ is transcribed into a noisy text by the pre-trained ASR model which is given as $\hat{t} = \mathrm{ASR}(a)$.
The user input is then constructed in the same format as during training: $u = u(\hat{t}, \mathcal{O}, \mathcal{K})$, where $\hat{t}$ is the ASR transcript, $\mathcal{O}=\{A,B,C,D\}$ denotes the multiple-choice options, and $\mathcal{K}$ is the KG context $(\mathcal{K}_{\text{sem}}, \mathcal{K}_{\text{phon}})$.  
The overall model input is given by $x = [s ; u]$, where $s$ is the fixed system instruction. The fine-tuned LLM generates the structured two-line output:
\begin{equation}
y = LLM_{\theta}(x) \;\Rightarrow\;
\left
\{
  \begin{array}{ll}
    \textit{Corrected Text:} & \tilde{t} \\
    \textit{Correct Option:} & o^*
  \end{array}
\right.
\label{eq:inference}
\end{equation}
with $\tilde{t}$ denoting the corrected transcript and $o^{*}\in\{A,B,C,D\}$ the predicted option.

%================================================================================
\section{Experiment and Results}
In this section, we first describe the data preparation and experimental settings. 
We then evaluate our MedSpeak framework on spoken medical QA against various baselines.

%=========================================================================
\subsection{Data Preparation}
To systematically evaluate MedSpeak, we use a diverse medical SQA benchmark by synthesizing spoken data from three well-established multiple-choice QA datasets \cite{zero-shot-sqa, medrag-healthcare, retrieval-augmented-dialogue}. The MMLU Medical Tasks dataset includes 1,089 questions covering six medical subjects, including biology, anatomy, and clinical medicine. The MedQA (USMLE-Based) dataset consists of 1,273 questions derived from the United States Medical Licensing Examination (USMLE), incorporating real-world diagnostic reasoning tasks. The MedMCQA dataset comprises 4,183 questions spanning multi-specialty medical assessments used in Indian medical entrance exams. Each dataset is synthesized at 16,000 Hz, mono WAV format using local pyttsx3 text-to-speech(TTS) engine for speech generation. This benchmark provides over 47 hours of spoken medical QA data, ensuring rigorous evaluation of ASR correction, retrieval-driven reasoning, and knowledge integration.

%=========================================================================
\subsection{Experimental Settings}

\textbf{Configurations:} Our MedSpeak framework is implemented by fine-tuning all parameters of Llama-3.1-8B-Instruct.
We conduct training across approximately $\sim$10k of vocalized medical QA pairs drawn from MedMCQA, MedQA, and MMLU-Medical, totaling about 6.79M  tokens. The model is optimized for 10 epochs with a batch size of 8 and gradient
the building up of 16, using a learning rate multiplier of 2 relative to the
cosine schedule baseline. The length of sequence is constrained to at most 2048 tokens, 
To prevent truncation of gold labels, we budget
%Budgeted 
semantic and phonetic knowledge-graph partitions of 600 and 300 tokens respectively.
% , are being assigned to prevent truncation of gold labels. 
The objective of training is the standard causal language modeling loss across the whole rendered dialogue, so that the model jointly learns ASR correction and answer prediction.
This experiment is conducted on a cluster of 8x  NVIDIA A100 GPUs with 80 GB memory each, paired with AMD EPYC 7742 64-Core Processor. We enable \texttt{bfloat16} precision, gradient checkpointing, and TF32 matrix multiplication for efficiency enhancement. The training runs are distributed across multiple GPUs with the help of the HuggingFace \texttt{Trainer} framework. 
The Whisper Small Model (244M parameters) is used as the front-end ASR system to generate the noisy transcript for all our experiments.

\noindent
\textbf{Baselines:}
We compare our MedSpeak framework against a set of baseline systems to evaluate the contribution of fine-tuning, ASR correction and KG integration:
(1) Zero-shot ASR (\textbf{ZS-ASR}) is the model using the raw whisper generated transcript without fine-tuning or KG-context.
(2) Zero-shot GT (\textbf{Zero-Shot}) is the model using the pure base LLM when provided with the ground truth transcripts, and without fine-tuning or KG-context.
(3) Fine-Tuned LLM + Whisper (\textbf{FT+Whisp}) is the model where the fine-tuned LLM is given the whisper generated transcripts without KG-context.
(4) Fine-Tuned LLM + GT (\textbf{FT-LLM}) is the fully fine-tuned LLM evaluated with the ground truth texts.
(5) \textbf{MedSpeak:} is our proposed model that incorporates noisy transcripts generated by  whisper with the budgeted KG context and the fined-tuned LLM. 

\noindent
\textbf{Evaluation Metrics:} 
We evaluate MedSpeak performance on  two metrics: (1) QA Accuracy (\%):  the fraction of correctly predicted multiple choice answers, and (2) Word Error Rate (WER) (\%): Punctuation-insensitive Levenshtein distance between reference and corrected transcripts, which quantifies the ASR correction quality,  calculated as: \(\text{WER} = \tfrac{S + D + I}{N} \times 100\%\), where $S$ are substitutions, $D$ deletions, $I$ insertions, and $N$ is the total number of words in the reference transcript.

%=========================================================================
\subsection{Results and  Analysis}
We compare the QA accuracy and WER of our proposed framework MedSpeak against different baselines for different datasets.
Accuracy results are listed in Table~\ref{tab:accuracy_results}. Indicatively, the best performance is recorded by MedSpeak overall on the benchmarks, recording an overall accuracy of 93.4\%, very close to the performance of the FT-LLM baseline paired with the ground-truth texts (92.5\%) and considerably higher than zero-shot setups in the 50-56\% range. The biggest boosts are recorded in Virology and in MedQA, where MedSpeak reduces the gap by over four percentage points compared to the FT-LLM baseline, demonstrating our knowledge graph context integration to be highly beneficial in coping with challenging terminology and noisy ASR transcriptions. The exception is in the case of MedMCQA where the LLM fine-tuned on ground-truth transcriptions performs by a very slim margin (92.5\% vs.\ 91.5\%). Even then though, MedSpeak outperforms baselines in all other tasks.
\begin{table}[htbp]
\centering
\vspace{-1em}
\caption{QA Accuracy of MedSpeak and Baselines.}
\label{tab:accuracy_results}
\scriptsize
\setlength{\tabcolsep}{2pt}
\renewcommand{\arraystretch}{0.9}
\begin{tabular}{lccccc}
\toprule
\textbf{Task} & \textbf{ZS-ASR} & \textbf{Zero-Shot} & \textbf{FT+Whisp} & \textbf{FT-LLM} & \textbf{MedSpeak} \\
\midrule
\multicolumn{6}{l}{\textbf{MMLU}} \\
Clinical    & 62.2 & 66.3 & 85.6 & 94.3 & 95.4 \\
Anatomy     & 57.0 & 64.4 & 85.2 & 94.1 & 93.3 \\
College     & 62.1 & 67.5 & 84.2 & 92.7 & 95.6 \\
Virology    & 47.6 & 48.8 & 85.5 & 91.0 & 95.8 \\
Prof. Med.  & 74.6 & 76.1 & 88.6 & 94.9 & 97.8 \\
\midrule
MedQA       & 54.9 & 58.9 & 86.6 & 91.8 & 97.5 \\
MedMCQA     & 45.5 & 52.8 & 82.3 & 92.5 & 91.5 \\
\midrule
\textbf{Avg.} & \textbf{50.2} & \textbf{56.3} & \textbf{83.7} & \textbf{92.5} & \textbf{93.4} \\
\bottomrule
\end{tabular}
\end{table}
Moreover, WER results are presented in Table~\ref{tab:wer_results}. As WER is defined only over ASR-processed inputs, ground-truth baselines are omitted; if used as input, their error rate would necessarily be zero. MedSpeak shows outright benefits. The overall WER decreases to 2.99\%, versus 7.72\% on ZS-ASR generated transcripts with non finetuned LLM and 3.58\% on FT+Whisp. Gains are particularly significant in Anatomy and Virology, where MedSpeak decreases error by 0.5–0.7 percentage points over the next-best baseline. These decreases are significant to how well combining semantic and phonetic knowledge graph information with fine-tuning helps the model to discriminate phonetically similar but semantically distinct medical entities and to correct domain-specific transcription mistakes with greater reliability. These results in tandem confirm that MedSpeak is equally very accurate on spoken medical QA and improves transcription quality.
\vspace{-1em}
\begin{table}[htbp]
\centering
\caption{WER(\%) of MedSpeak and Baselines.}
\label{tab:wer_results}
\scriptsize
\setlength{\tabcolsep}{3pt}
\renewcommand{\arraystretch}{0.9}
\begin{tabular}{lccc}
\toprule
\textbf{Task} & \textbf{ZS-ASR} & \textbf{FT+Whisp} & \textbf{MedSpeak} \\
\midrule
\multicolumn{4}{l}{
    \textbf{MMLU}
} \\
Clinical    & 7.67 & 2.73 & 2.21 \\
Anatomy     & 8.70 & 2.58 & 1.98 \\
College     & 6.75 & 2.10 & 1.57 \\
Virology    & 7.89 & 2.46 & 1.75 \\
Prof. Med.  & 6.19 & 3.93 & 3.78 \\
\midrule
MedQA       & 6.39 & 4.49 & 4.35 \\
MedMCQA     & 8.14 & 3.54 & 2.81 \\
\midrule
\textbf{Avg.} & \textbf{7.72} & \textbf{3.58} & \textbf{2.99} \\
\bottomrule
\end{tabular}
\vspace{-2em}
\end{table}

%=================================================================
\section{Conclusion}
 In this paper, we proposed MedSpeak, a framework integrates semantic and phonetic context from a medical knowledge graph with the reasoning ability of LLMs, enabling robust correction of transcription errors and reliable clinical reasoning.
 % A multi-constraint retrieval process was developed to align phonetic similarity and semantic relations with ASR hypotheses, while a fine-tuned LLM unifies error correction and answer generation within a single objective. 
Through extensive evaluation, MedSpeak demonstrated consistent improvements in both transcription accuracy and downstream QA reliability, 
offering a scalable and effective solution for enhancing ASR systems in high-stakes medical applications.

\newpage
\bibliographystyle{IEEEbib}
\bibliography{strings}

\end{document}